\documentclass[10pt,twocolumn,letterpaper]{article}

\usepackage{cvpr}              

\usepackage{multirow}
\usepackage[table]{xcolor}
\definecolor{lightgreen}{RGB}{240,255,240}
\definecolor{lightblue}{RGB}{240,240,255}
\definecolor{cvprblue}{rgb}{0.21,0.49,0.74}
\usepackage[pagebackref,breaklinks,colorlinks,allcolors=cvprblue]{hyperref}

\def\paperID{4197} 
\def\confName{CVPR}
\def\confYear{2026}

\title{LoG3D: Ultra-High-Resolution 3D Shape Modeling via Local-to-Global Partitioning}

\author{Xinran Yang$^{1,2\dag}$, Shuichang Lai$^{2\dag}$, Jiangjing Lyu$^{2*}$, Hongjie Li$^2$, Bowen Pan$^2$, \\ Yuanqi Li$^1$, Jie Guo$^1$, Zhengkang Zhou$^3$, Yanwen Guo$^{1*}$\\
$^{1}$Nanjing University  $^{2}$Alibaba
Group $^{3}$Nanjing Urban Construction Tunnel\&Bridge Intelligent Management\\
}

\begin{document}
\twocolumn[{
\renewcommand\twocolumn[1][]{#1}
\maketitle

\vspace{-5mm}
\begin{center}
    \captionsetup{type=figure}
    \includegraphics[width=\textwidth]{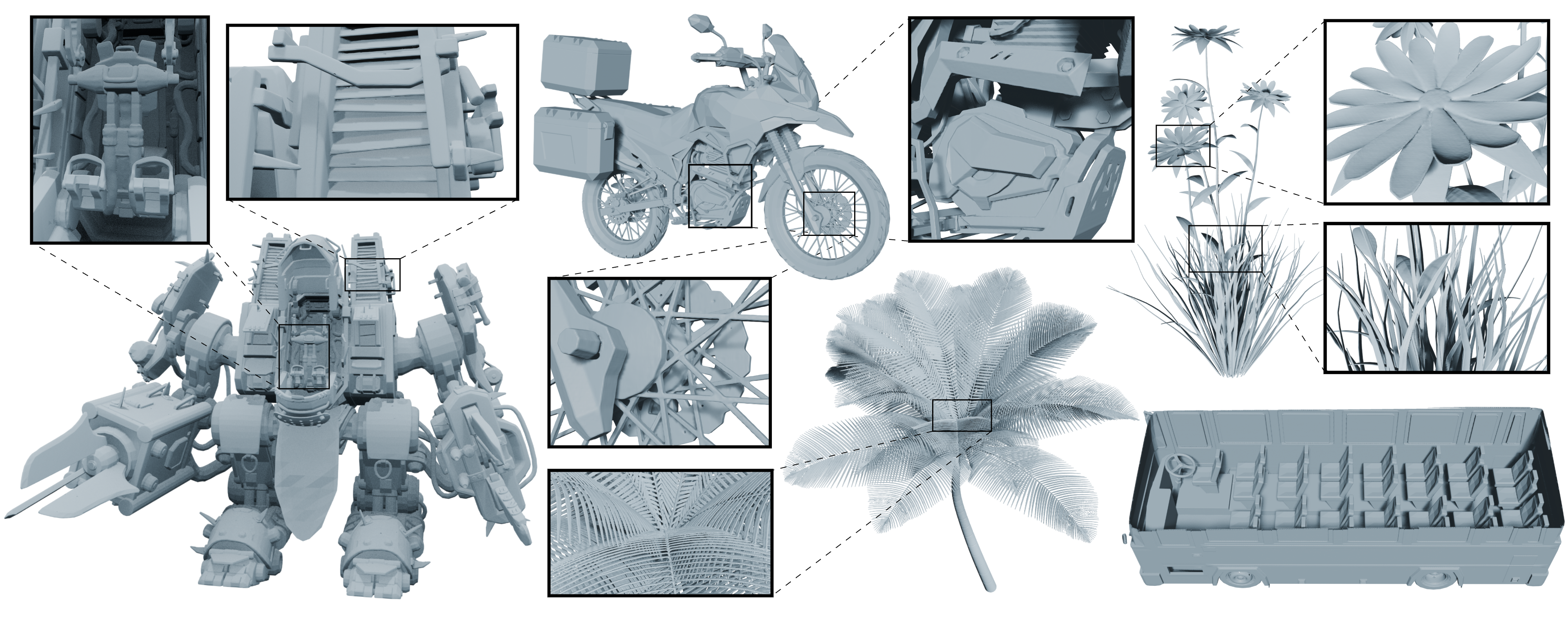}
 \vspace{-5mm}
  \caption{\textbf{LoG3D Reconstruction Results.} 
  LoG3D sets a new state of the art in 3D reconstruction, delivering high-fidelity results on challenging inputs at resolution up to $2048^3$. The model demonstrates remarkable robustness, adeptly handling non-watertight surfaces, preserving internal structural integrity, and faithfully reconstructing highly complex geometries. Best viewed with zoom-in for fine details.}
  \label{fig:teaser}
\end{center}
}]
\renewcommand{\thefootnote}{} 
\footnotetext{$\dag$Equal contribution. $^*$Corresponding authors.}

\begin{abstract}

Generating high-fidelity 3D contents remains a fundamental challenge due to the complexity of representing arbitrary topologies—such as open surfaces and intricate internal structures—while preserving geometric details. Prevailing methods based on signed distance fields (SDFs) are hampered by costly watertight preprocessing and struggle with non-manifold geometries, while point-cloud representations often suffer from sampling artifacts and surface discontinuities. To overcome these limitations, we propose a novel 3D variational autoencoder (VAE) framework built upon unsigned distance fields (UDFs)—a more robust and computationally efficient representation that naturally handles complex and incomplete shapes. Our core innovation is a local-to-global (LoG) architecture that processes the UDF by partitioning it into uniform subvolumes, termed UBlocks. This architecture couples 3D convolutions for capturing local detail with sparse transformers for enforcing global coherence. A Pad-Average strategy further ensures smooth transitions at subvolume boundaries during reconstruction. This modular design enables seamless scaling to ultra-high resolutions up to $2048^3$-a regime previously unattainable for 3D VAEs. Experiments demonstrate state-of-the-art performance in both reconstruction accuracy and generative quality, yielding superior surface smoothness and geometric flexibility.

\end{abstract}    
\section{Introduction}
\label{sec:intro}

The field of 3D generative AI has witnessed remarkable progress~\cite{wang2023rodin, tochilkin2024triposr, hunyuan3d2025hunyuan3d, xiang2025structured}, driving novel applications across diverse domains such as entertainment~\cite{chen2024rapid, li2023synthesizing, li2024advances}, design, and robotics~\cite{yang2024holodeck, yang2024physcene}. Despite this progress, generating high-fidelity 3D contents remains a far more formidable challenge than its 2D counterparts, a difficulty rooted in the inherent complexity of representing and processing 3D geometry. This difficulty is especially acute when striving to achieve both high geometric fidelity and support for arbitrary topologies, such as open surfaces and intricate internal structures. 

To tackle this challenge, prevailing models have explored a diverse spectrum of 3D representations, including point clouds~\cite{luo2021diffusion, he2025sparseflex, melas2023pc2, nichol2022point, chen2025dora}, meshes~\cite{siddiqui2024meshgpt, nash2020polygen, chen2024meshanything, wang2024llamameshunifying3dmesh}, 3D Gaussian Splatting (3DGS)~\cite{zhang2024gaussiancube, yushi2025gaussiananything, xiang2025structured}, and implicit fields~\cite{cheng2023sdfusion, li2025craftsman3d, ren2024xcube, zhang20233dshape2vecset, zhang2024clay, zhao2023michelangelo, zheng2023locally}. Among these, implicit fields like Signed Distance Fields (SDFs) excel at modeling continuous surfaces~\cite{wu2025direct3d, wu2024direct3ds1, zhang2024clay}. However, their reliance on costly watertight preprocessing severely limits their applicability; this step discards geometric details and fundamentally fails on open or non-manifold topologies. Alternatively, point clouds circumvent the constraint of watertightness by representing geometry explicitly ~\cite{he2025sparseflex, melas2023pc2, chen2025dora, hunyuan3d2025hunyuan3d}. Nevertheless, this simplicity comes at a price: as a discrete representation, they are sensitive to sampling density and often produce surfaces plagued by holes and discontinuities.

In this paper, we pivot to Unsigned Distance Fields (UDFs)  as our core representation, chosen for their superior efficiency, robustness, and topological flexibility compared to alternatives like SDFs and point clouds. As a continuous field, a UDF is a more natural fit for neural networks than discrete point clouds. Critically, by bypassing the costly and error-prone sign computation of SDFs, our UDF-based approach is not only more efficient but also inherently robust to noise and imperfections in real-world data. This topology-agnostic nature also allows it to faithfully represent complex, non-manifold geometries where SDFs are ill-posed or computationally intractable.

However, the adoption of efficient and flexible UDFs still faces a significant challenge when integrated with existing VAE architectures. Typically, sparse voxel-based neural models \cite{xiang2025structured,wu2025direct3d, he2025sparseflex,li2025sparc3d} rely on an aggressive compression-decompression pipeline: the entire input is downsampled into a latent representation and then upsampled for reconstruction. This global-bottleneck design inherently risks losing high-frequency geometric details, which are crucial for fidelity. To circumvent this, we introduce a novel local-to-global (LoG) architecture combining the strengths of 3D convolutions and transformers \cite{vaswani2017attention, liu2021swin}. Our approach begins by partitioning a high-resolution UDF into a grid of uniformly sized subvolumes (termed UBlocks). This strategy avoids premature global compression. Instead, a lightweight 3D CNN operates within each UBlock to preserve fine-grained local geometry. These locally-refined feature blocks are then processed by sparse transformers, which models their long-range dependencies to capture global structural coherence. To ensure seamless continuity across UBlock boundaries during reassembly, we propose a simple yet effective Pad-Average strategy that minimizes reconstruction artifacts.

This hierarchical design is not only robust to varying input scales but also scales effortlessly to ultra-high resolutions. Our framework achieves state-of-the-art performance in 3D geometry reconstruction, delivering high-fidelity shapes with notably smoother surfaces at ultra-high resolutions up to $2048^3$—a regime previously unattainable for existing VAE-based models \cite{xiang2025structured, chen2025dora, hunyuan3d2025hunyuan3d, wu2025direct3d, he2025sparseflex, li2025sparc3d}.

In summary, our main contributions are:
\begin{itemize}
\item We propose \textbf{UBlock}, a novel representation partitioning a high-resolution UDF into uniform subvolumes. This design decouples representation resolution from model complexity, enabling unprecedented scalability to ultra-high resolutions (e.g., $2048^3$).

\item We introduce \textbf{LoG-VAE}, a modality-consistent variational autoencoder that integrates local 3D convolutions with global sparse transformers. This hierarchical approach processes UBlocks efficiently, preserving fine-grained details while avoiding the information loss common in global-bottleneck designs.


\item Our framework achieves state-of-the-art results in both reconstruction and generation. It produces smoother, more topologically complex shapes than all existing methods, setting a new standard for VAE-based 3D models.
\end{itemize}

\section{Related Work}

\subsection{3D Shape Representations for Generation}

\textbf{Explicit 3D Representations.}
Triangle meshes and point clouds are the most common explicit representations of 3D data. Triangle meshes, defined by vertices and faces, are celebrated for their ability to capture high-fidelity surface details and arbitrary topologies. However, their inherent non-Euclidean graph structure poses a significant challenge for deep learning frameworks, which must contend with irregular connectivity and the lack of a canonical ordering. While recent autoregressive models \cite{chen2024meshanything, chen2025meshanything, siddiqui2024meshgpt, wang2024llamameshunifying3dmesh, hao2024meshtron} attempt to tackle this by sequentially generating geometry and connectivity, they are often bottlenecked by prohibitive sampling times and limited long-range context. 
As a structurally simpler alternative, point clouds represent geometry as an unordered set of 3D points. This simplicity makes them highly amenable to modern generative architectures~\cite{luo2021diffusion, he2025sparseflex}. Yet, this simplicity comes at a cost: the representation discards all explicit connectivity information. Consequently, learning to reconstruct a continuous manifold is a challenging task in itself. The process is highly sensitive to sampling density, and the resulting surfaces are often plagued by holes and other topological artifacts.

\noindent \textbf{Implicit 3D Representations.}
Implicit representations are widely used in geometry learning, particularly for 3D reconstruction~\cite{guillard2022meshudf, wang2021neus, huang2023neural, chen20223psdf} and generation. Among these, Signed Distance Fields (SDFs) are a dominant paradigm, defining geometry as the zero-level set and achieving state-of-the-art generation results~\cite{wu2025direct3d, wu2024direct3ds1, li2025sparc3d, ren2024xcube}. However, this SDF-based method is fundamentally constrained by its reliance on a consistent inside/outside sign. This necessitates a costly, often lossy, preprocessing step to convert input meshes into watertight manifolds, precluding the representation of common open or non-manifold geometries.
 While Unsigned Distance Fields (UDFs) inherently avoid this limitation, they have received comparatively less attention for generative modeling, with existing attempts often struggling to capture fine-grained details or maintain global coherence, especially at high resolutions~\cite{yu2024surf}. This creates a clear need for a robust and scalable framework that leverages UDFs for arbitrary topologies—a challenge our work directly addresses.

\subsection{3D Shape Variational Auto-Encoders}

\textbf{VecSet-based VAEs.}
VecSet-based VAEs have emerged as a prominent approach, seeking to bridge local geometric detail with a structured global latent space. The foundational 3DShape2VecSet~\cite{zhang20233dshape2vecset} established this paradigm by encoding local point features into a set of latent vectors using a transformer. Subsequent research has refined this architecture, focusing on scalability (CLAY~\cite{zhang2024clay}), expressiveness through Mixture-of-Experts (TripoSG~\cite{li2025triposg}), and more efficient sampling strategies (Dora~\cite{chen2025dora}, Hunyuan2~\cite{hunyuan3d2025hunyuan3d}).

Despite these advances, a fundamental architectural flaw persists across all such models: a modality mismatch between the input and output. They compress discrete, local point features into a global latent set, only to then decode these vectors back into a continuous local field (e.g., SDF values). This forces the VAE to be burdened with a dual mandate: it must perform both high-level semantic abstraction and low-level modality conversion. This conflation of responsibilities places an immense burden on the attention mechanism and necessitates increasingly complex and parameter-heavy models to resolve the mismatch.

\noindent \textbf{Sparse voxel–based VAEs.}
To mitigate the spatial information loss inherent in VecSet-based VAEs, sparse voxel-based VAEs represent shapes using features anchored to a sparse voxel grid. Pioneering this direction, XCube~\cite{ren2024xcube} attached local geometric features (e.g., SDF values and normals) to voxels, demonstrating improved detail preservation. This concept was extended by TRELLIS~\cite{xiang2025structured}, which incorporated semantic DINOv2 features~\cite{oquab2023dinov2} for joint modeling, and further scaled by TripoSF~\cite{he2025sparseflex} for high-resolution reconstruction. Despite preserving spatial structure, these methods still suffer from the modality mismatch between abstract input features and the target metric SDF field. Acknowledging this bottleneck, recent works like Sparc3D~\cite{li2025sparc3d} and Direct3D-S2~\cite{wu2025direct3d} adopt a more direct approach, using the SDF field itself as both input representation and reconstruction target. This strategy successfully eliminates the modality conversion step and has demonstrated superior reconstruction quality. However, by fully committing to an SDF-centric pipeline, these models inherit the very limitation we identified earlier: the strict requirement for watertight input meshes. This motivates our work: a UDF-based VAE designed to handle arbitrary topologies without the limitations of an SDF-centric pipeline.
\section{Method}

\begin{figure*}[h]
  \vspace{-3mm}
  \centering
  \includegraphics[width=\linewidth]{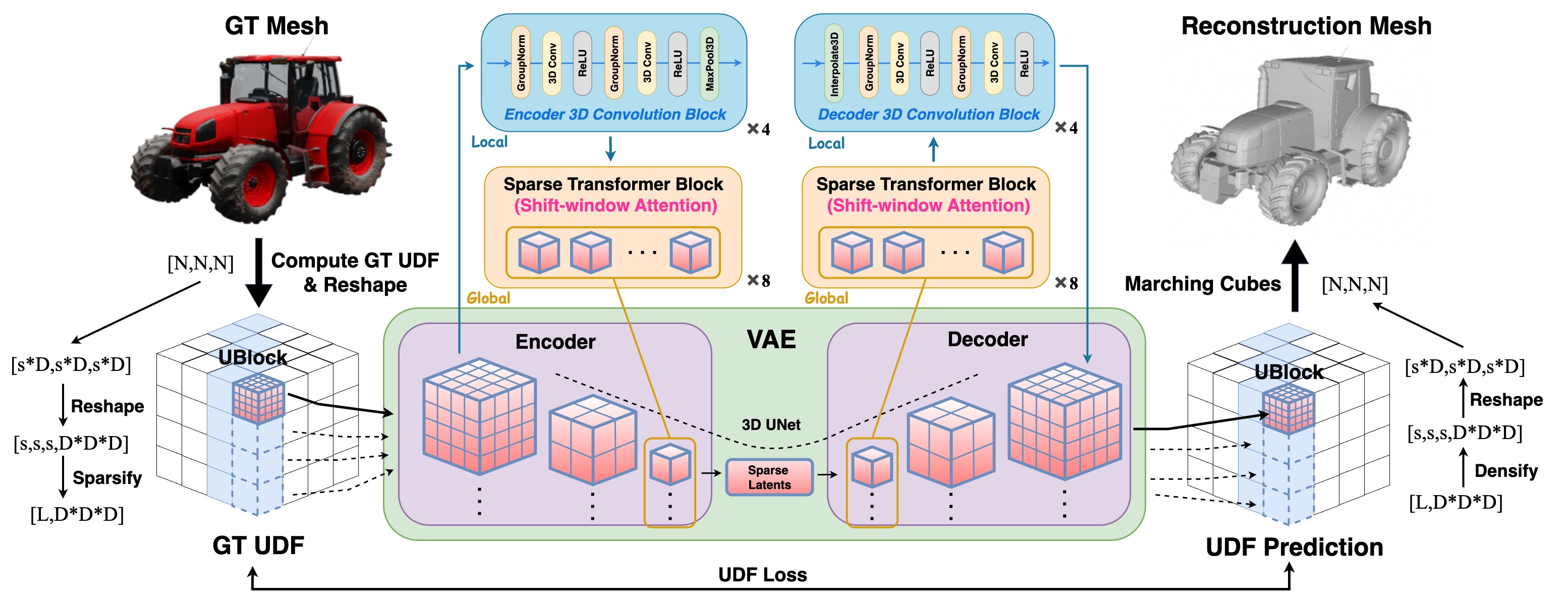}
  \caption{\textbf{Pipeline of the LoG-VAE Framework.} An input UDF is partitioned into sparse UBlocks, which are encoded into latent codes by a local-global encoder (3D Conv and sparse transformers). A symmetric decoder reconstructs the UBlocks, which are then reassembled into a full UDF field for final mesh extraction. The entire network is trained under the supervision of the UDF loss.}
  \label{fig:Pipeline}
\end{figure*}



\subsection{Overview}
We present LoG-VAE, a novel local-to-global variational autoencoder for ultra-high-resolution 3D shape modeling. Our approach operates on Unsigned Distance Fields (UDFs), introducing a partitioning scheme that enables unprecedented scalability for generative modeling at resolutions up to $2048^3$. As illustrated in \cref{fig:Pipeline}, the process begins by decomposing the input UDF field into a uniform grid of padded subvolumes (termed UBlocks). A hybrid encoder, which synergistically combines local 3D convolutions with global sparse transformers, then maps these UBlocks into a compact set of sparse latent vectors. Subsequently, a symmetric decoder reconstructs the UBlock features from these latents. These decoded blocks are seamlessly reassembled into a complete UDF field via our proposed pad-average strategy. Finally, the high-resolution mesh is extracted from this field using the Marching Cubes algorithm~\cite{lorensen1998marching}.

\subsection{UDF Voxel Input}
Our method begins with a triangle mesh $\mathcal{M}$, which is converted into an Unsigned Distance Field (UDF) and discretized into a dense volume $\mathcal{V} \in \mathbb{R}^{N \times N \times N}$. To manage computational costs at high resolutions, we create a sparse representation $\mathcal{V}_{sparse}$ by retaining only voxels near the surface whose distance values $s(\mathbf{x})$ are below a threshold $\tau$:
\begin{equation}
   \mathcal{V}_{sparse} = \left\{(\mathbf{x}_i, u(\mathbf{x}_i)) \mid u(\mathbf{x}_i) < \tau \right\}_{i=1}^{|\mathcal{V}_{sparse}|}.
\end{equation}
We set $\tau = 4/N$ and clip all other distance values to a maximum of $5/N$. The entire volume is then min-max normalized to $[0, 1]$. 

\noindent \textbf{UBlock Partitioning.} 
The dense volume $\mathcal{V}$ is first partitioned into a grid of $D \times D \times D$ subvolumes, termed UBlocks, where the subvolume resolution $D$ is set by a partition factor $s$ ($D=N/s$). From this complete grid, we only select the $L$ ``active" UBlocks that contain voxels from the sparse set $\mathcal{V}_{sparse}$. The UBlock is defined as:
\begin{equation}
\mathcal{U} = \{(\mathbf{f}_i, \mathbf{p}_i)\}_{i=1}^L
\end{equation}
where $\mathbf{f}_i \in \mathbb{R}^{D \times D \times D}$ is the feature tensor containing the normalized UDF values for the $i$-th block, $\mathbf{p}_i \in \mathbb{Z}^3$ is its subvolume grid coordinate, and for typical sparse shapes, $L \ll s^3$. 
This partitioning scheme is the cornerstone of our approach, as it decouples model complexity from the input resolution $N$. By operating on fixed-size UBlocks, our model can preserve high-frequency details while remaining scalable. The fidelity of the final reassembled UDF is directly tied to the local reconstruction quality of each UBlock. This design inherently supports arbitrary input resolutions without architectural changes, enabling seamless scaling up to $2048^3$.

\subsection{Network Architecture}

The preceding processing stage transforms an input mesh $\mathcal{M}$ into a set of UBlocks $\mathcal{U}$. These UBlocks are then passed through our model, the \textbf{LoG-VAE}, where an encoder $\mathbf{E}$ maps these to a sparse, structured latent set $\mathcal{Z}_{slat}$, and a corresponding decoder $\mathbf{D}$ reconstructs them as $\hat{\mathcal{U}}$. The final mesh $\hat{\mathcal{M}}$ is then extracted from the reassembled blocks. This process can be formulated as:
\begin{equation}
    \mathcal{V} = \textbf{UDF}(\mathcal{M});
    \mathcal{U} = \textbf{Partition}(\mathcal{V})
\end{equation}
\begin{equation}
    \mathcal{Z}_{slat} = \mathbf{E}(\mathcal{U}); \quad \hat{\mathcal{U}} = \mathbf{D}(\mathcal{Z}_{slat})
\end{equation}
\begin{equation}
    \hat{\mathcal{M}} = \textbf{MeshExtract}(\hat{\mathcal{U}}).
\end{equation}

As illustrated in \cref{fig:Pipeline}, our network features a symmetric encoder-decoder architecture. In our VAE, local 3D convolutions extract spatial features from each UBlock, while global sparse transformers facilitates inter-block communication through shift-window attention mechanisms. This hybrid design effectively combines fine-grained local feature extraction with global structural reasoning.


\noindent \textbf{Encoder.} The encoder $\mathbf{E}$ employs a hybrid framework combining local 3D convolution networks and global sparse 3D transformer networks. We first extract local geometric features within UBlocks by 3D convolution with 3D maxpooling operations, progressively downsampling the spatial resolution. We then process the sparse UBlocks as variable-length tokens and utilize shifted window attention to capture local contextual information between the valid UBlocks. Inspired by TRELLIS~\cite{xiang2025structured}, the feature of each valid voxel is augmented with positional encodding based on its 3D coordinates before being fed into 3D shift window attention layers. 

\noindent \textbf{Decoder.} The decoder $\mathbf{D}$ adopts a symmetric structure with respect to the encoder, leveraging global sparse attention layers and local 3D CNN blocks to progressively upsample the latent representation $\mathcal{Z}_{slat}$ to $\hat{\mathcal{U}}$. To reconstruct the final mesh, we first map the processed $\hat{\mathcal{U}}$ features back to their original spatial positions in the UDF volume of resolution $N^3$. Then, we apply inverse normalization to restore the unsigned distance values to their original dynamic range. Since UDF values are always larger than zero, we apply Marching Cubes~\cite{lorensen1998marching} algorithm with isosurface threshold $\theta=1/N$(where N denotes the UDF resolution) to extract the final 3D mesh reconstruction $\hat{\mathcal{M}}$. 

\noindent \textbf{Training Losses.} The decoded number of sparse UBlocks is the same as input UBlocks, as $|\hat{\mathcal{U}}|=|\mathcal{U}|=L$. We enforce supervision on the UDF values across all these spatial positions in all UBlocks. The term of KL-divergence regularization is imposed on the latent representation $\mathcal{Z}_{slat}$ to constrain excessive variations the the latent space. The overall training objective of our model is formulated as:
\begin{equation}
    \mathcal{L}_{udf} = \frac{1}{|\hat{\mathcal{U}}|} \sum_{(\mathbf{x},\hat{u}(\mathbf{x})) \in \hat{\mathcal{U}}} \lVert u(\mathbf{x}) - \hat{u}(\mathbf{x}) \rVert^{2}_{2}
    \label{eq:loss_udf}
\end{equation}

\begin{equation}
    \mathcal{L}_{total} = \mathcal{L}_{udf} + \lambda \mathcal{L}_{KL}
\end{equation}
where $\lambda$ denote the weight of KL-divergence loss. In our experiments, we substitute the standard L2 loss with the Huber loss for the reconstruction term (\cref{eq:loss_udf}), as it is less sensitive to outliers and thus provides greater training stability.

\subsection{Pad-Average Strategy}
\begin{figure}[h]
  \centering
  \includegraphics[width=\linewidth]{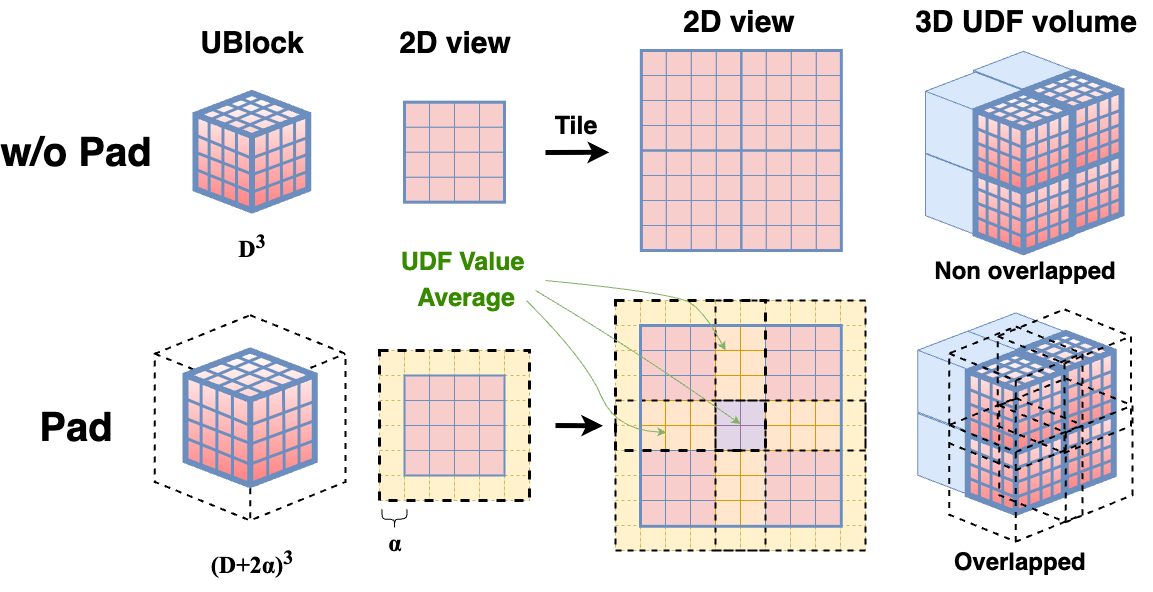}
\caption{\textbf{Illustration for our Pad-Average strategy.}}
  \label{fig:PadAverage}
\end{figure}

Direct reassembling UBlocks to the global UDF volume can introduce geometric discontinuities at block boundaries due to the lack of spatial coherence constraints. These artifacts manifest as surface roughness and topological inconsistencies in the reconstructed mesh, particularly at the interfaces between adjacent UBlocks with disparate feature representations. To address this, we introduce a Pad-Average strategy for our LoG-VAE framework. 

As shown in \cref{fig:PadAverage}, for each UBlock, we employ padded volumes to get expanded Ublock resolution $(D+2\alpha)^3$ (where $\alpha$ is the padding size) as network inputs, rather than the original $D^3$ UBlock resolution. The total block count $L$ remains constant across the transformation; however, adjacent UBlocks now exhibit spatial overlap to enhance inter-block coherence. These padded UBlocks are fed into our encoder, transformed into latents and then recovered by our decoder. 
During the reassembly stage, we map these overlapping UBlocks to the $N^3$ UDF volume and compute the average of the values in overlapping regions to determine the final UDF value. The smoothly reconstructed UDF field is then used to generate the final mesh via the Marching Cubes algorithm~\cite{lorensen1998marching}. This Pad-Average strategy effectively mitigates boundary artifacts while preserving geometric fidelity, yielding smoother extracted surfaces.

\section{Experiments}
\label{Experiments}
\subsection{Implementation Details}
We implement our model on top of the official TRELLIS~\cite{xiang2025structured} codebase. The model is trained on a curated dataset of approximately 500,000 high-quality 3D meshes, selected through rigorous quality filtering from large-scale repositories including ABO~\cite{collins2022abo}, HSSD~\cite{khanna2024habitat}, and ObjaverseXL~\cite{deitke2023objaverse, deitke2023objaversexl}. During pre-processing, we compute UDF representations using an adaptive threshold of $\tau=4/N$ (where N is the target UDF resolution) and apply padding to build UBlocks as sparse tensor input. 

The LoG-VAE framework is first trained on $1024^3$ UDFs (Ours-1024, $s=128$, $D=8$), and then fine-tuned to support $2048^3$ UDFs (Ours-2048, $s=256$, $D=8$), enabling progressive learning of geometric details across scales. We set the default padding value $\alpha=2$ and the latent code channel to $16$. The model was trained for five days on 8 NVIDIA H20 GPUs, using a batch size of 1. We employ the AdamW~\cite{loshchilov2017decoupled} optimizer with an initial learning rate of $5\times10^{-5}$. At inference, we generate the results with Marching Cubes \cite{lorensen1998marching}. More implementation details can be found in the supplementary material.

  
\subsection{Dataset, Baselines and Metrics}
\noindent \textbf{Dataset.}
To comprehensively evaluate and compare the reconstruction quality of various VAE methods, we construct two distinct test sets. The first is a curated subset of the Toys4K benchmark~\cite{stojanov2021using}, comprising models with high-frequency geometric details. The second is our in-house \textit{iHome} dataset, which consists of common household items. The object categories in \textit{iHome} deliberately differ from our training distribution, allowing us to assess out-of-distribution generalization ability of our model. 

\noindent \textbf{Baselines.}
We compare our VAE with state-of-the-art methods, including Hunyuan3D-2.1~\cite{hunyuan3d2025hunyuan3d} (resolution $256^3$), TRELLIS~\cite{xiang2025structured} (resolution $256^3$), Dora~\cite{chen2025dora} (resolution $256^3$), Direct3D-S2~\cite{wu2025direct3d}  (resolution $1024^3$) and TripoSF~\cite{he2025sparseflex} (resolution $1024^3$). We use the officially provided pre-trained models for all baselines. For the conditional generation task, we compare our method against Direct3D-S2~\cite{wu2025direct3d} and Sparc3D~\cite{li2025sparc3d}.

\noindent \textbf{Metrics.}
We evaluate the reconstruction performance of our VAE using several standard metrics. For overall geometric accuracy, we report Chamfer Distance (CD) and F-score. The F-score is calculated at two thresholds: 0.01 (F1-0.01) and 0.001 (F1-0.001).
To evaluate fine details, we adopt two further metrics from Dora~\cite{chen2025dora}: a multi-view Normal Mean Squared Error (NMSE), which compares rendered normal maps against the ground truth, and the Sharp Normal Error (SNE), which specifically measures reconstruction quality in salient regions and along sharp edges.

\begin{figure*}
\vspace{-8mm}
  \centering
\includegraphics[width=0.95\linewidth]{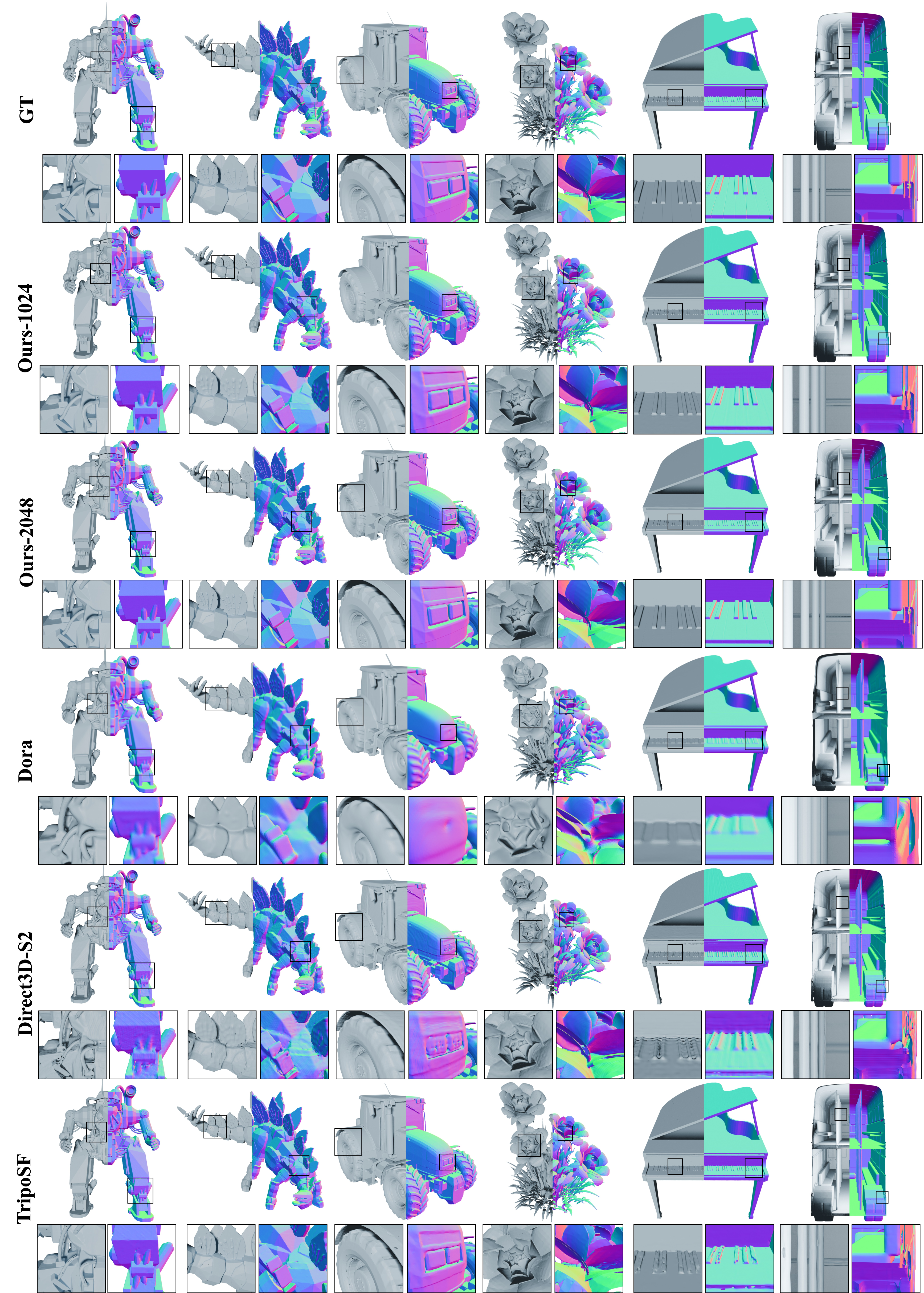}
\vspace{-2mm}
  \caption{\textbf{Qualitative comparison of VAE reconstruction.} Our approach demonstrates superior performance in reconstructing complex shapes, open surfaces, and even interior structures. Best viewed with zoom-in.}
  \label{fig:comparisons}
\end{figure*}

\subsection{VAE Reconstruction Evaluation}

\begin{table*}[h]
  \centering
  \caption{\textbf{Quantitative comparison on the 
  Toys4k~\cite{stojanov2021using} and iHome datasets.} Chamfer Distance (CD, $\times 10^5$) and F1 score (F1, $\times 10^2$) are reported.}
  \resizebox{\linewidth}{!}{
  \begin{tabular}{l|ccccc|ccccc}
    \toprule
    \multirow{2}{*}{
    \textbf{Method}} & \multicolumn{5}{c|}{Toys4k~\cite{stojanov2021using}} & \multicolumn{5}{|c}{iHome} \\

    \cmidrule(r){2-11}
    & NMSE $\downarrow$ & SNE $\downarrow$ & CD $\downarrow$ & F1(0.001) $\uparrow$ & F1(0.01) $\uparrow$ &  NMSE $\downarrow$ & SNE $\downarrow$ & CD $\downarrow$ & F1(0.001) $\uparrow$ & F1(0.01) $\uparrow$\\
    
    \midrule
    Hunyuan3D-2.1 \cite{hunyuan3d2025hunyuan3d} & 3.00 & 18.74 & 0.54 & 7.22 & 98.77 & 1.31 & 20.98 & 0.21 & 5.73 & 99.88\\
    TRELLIS \cite{xiang2025structured} & 3.30 & 14.51 & 0.39 & 20.01 & 98.50 & 3.47 & 23.05 & 0.16 & 11.80 & 99.68 \\
    Dora \cite{chen2025dora} & 4.66 & 21.93 & 0.39 & 6.18 & 99.68 & 1.49 & 20.10 & 0.21 & 5.31 & 99.86 \\
    Direct3D-S2 \cite{wu2025direct3d} & 3.17 & 12.35 & 0.23 & 21.99 & 99.81 & 
    1.21 & 15.99 & 0.12 & 11.39 & 99.91 \\
    TripoSF \cite{he2025sparseflex} & 1.27 & 6.38 & 0.07 & 36.16 & 99.94 & 0.88 & 8.63 & \textbf{0.06} & 30.99 & 99.95\\
    \midrule
    \rowcolor{lightgreen}
    Ours-1024 & 0.34 & 1.13 & \textbf{0.06} & 42.85 & 99.94 & 0.18 & 1.19 & \textbf{0.06} & 38.52 & 99.95\\
    \rowcolor{lightblue}
    Ours-2048 & \textbf{0.29} & \textbf{0.85} & \textbf{0.06} & \textbf{42.98} & \textbf{99.96} & \textbf{0.17} & \textbf{0.94} & \textbf{0.06} & \textbf{39.37} & \textbf{99.96} \\
    \bottomrule
  \end{tabular}
  }
  \label{tab:comparisons}
\end{table*}

\noindent \textbf{Quantitative Comparison.} We conduct extensive experiments to evaluate the quantitative results of VAE reconstruction from different methods on the Toys4k and iHome dataset in \cref{tab:comparisons}. Our VAE model operates at the same compression ratio as Direct3D~\cite{wu2025direct3d} and achieves a four-fold higher compression rate compared to TripoSF~\cite{he2025sparseflex}.
Our model, even at a $1024^3$ resolution, consistently outperforms all baselines across every metric on both datasets. When scaled to $2048^3$, its performance improves further, validating the exceptional scalability and fidelity of our approach. This superior performance is a direct result of the LoG-VAE design. By leveraging UBlocks, our model effectively performs a high-fidelity compression of local geometric details. A key advantage of our architecture is its decoupling of model size from input resolution; scaling to higher resolutions like $2048^3$ only increases the number of sparse tokens ($L$) without requiring changes to the model parameters. Consequently, the final reconstruction quality scales naturally with the increased resolution of the input data.

\noindent \textbf{Qualitative Comparison.} \cref{fig:comparisons} qualitatively evaluates the superiority of our method, particularly on complex shapes with fine-grained details, open surfaces, and interior structures. Our method avoids surface artifacts like the small holes common in TripoSF's~\cite{he2025sparseflex} results. It simultaneously preserves the smoothness of large surfaces, captures fine geometric details, and maintains the integrity of internal structures, achieving a final quality highly faithful to the ground truth. The effectiveness of our method stems from the interplay of its core components. The local 3D convolutions within each UBlock are responsible for the high-fidelity compression and restoration of local spatial details. Meanwhile, the global sparse Transformer ensures overall coherence by modeling the relationships between these blocks. Finally, the Pad-Average strategy enforces smoothness across UBlock boundaries, resulting in a continuous and seamless final surface. More qualitative comparison with Hunyuan3D-2.1~\cite{hunyuan3d2025hunyuan3d} and TRELLIS~\cite{xiang2025structured} can be found in the supplymentary material.


\subsection{Image to 3D Generation}
\begin{figure}
  \centering
\includegraphics[width=\linewidth]{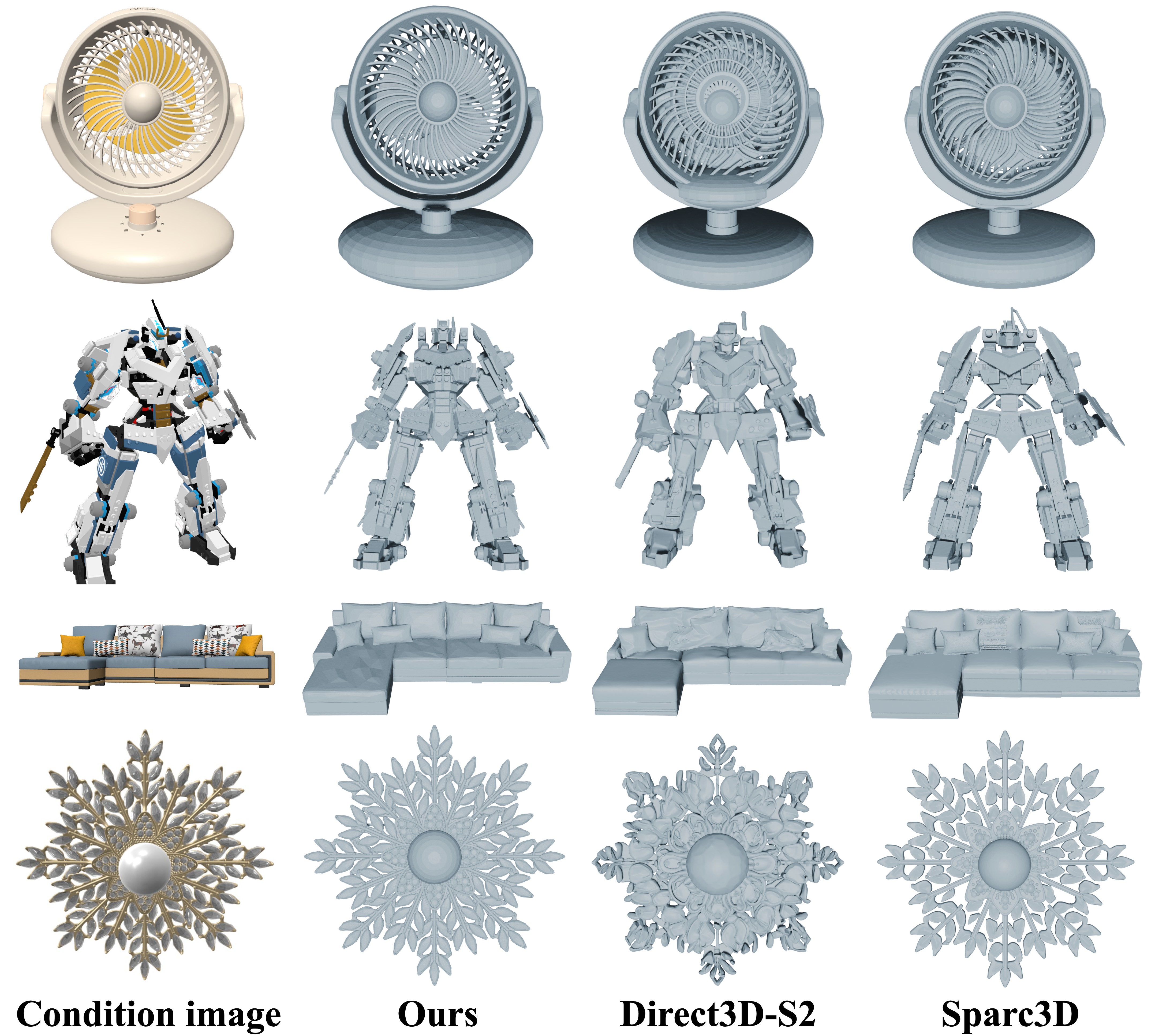}
  \caption{\textbf{Qualitative comparison of simgle-image-to-3D generation}. When comparing reconstructions at a $1024^3$ resolution, the generator trained with our LoG-VAE yields more detailed reconstructions than Direct3D-S2 \cite{wu2025direct3d} and Sparc3D \cite{li2025sparc3d}.}
  \label{fig:generation}
  \vspace{-2mm}
\end{figure}

To demonstrate its efficacy as a generative prior, we integrate LoG-VAE into a conditional generation pipeline. Following TRELLIS~\cite{xiang2025structured}, we first train a Structure Flow Model to predict a sparse $128^3$ shape structure from an input image. We then adapt their Structured Latent Flow Model to leverage our powerful decoder for high-resolution generation. As showcased in \cref{fig:generation}, this pipeline generates high-fidelity shapes with exceptional detail. Our model excels at preserving sharp edges, maintaining surface smoothness, and rendering intricate geometries, exemplified by the cleanly defined fan blades and unbroken sofa cushions. The strong alignment between the generated 3D models and the input images underscores the robust generalization capabilities of our learned latent space.

\subsection{Ablation Studies}
\begin{figure}
  \centering
\includegraphics[width=\linewidth]{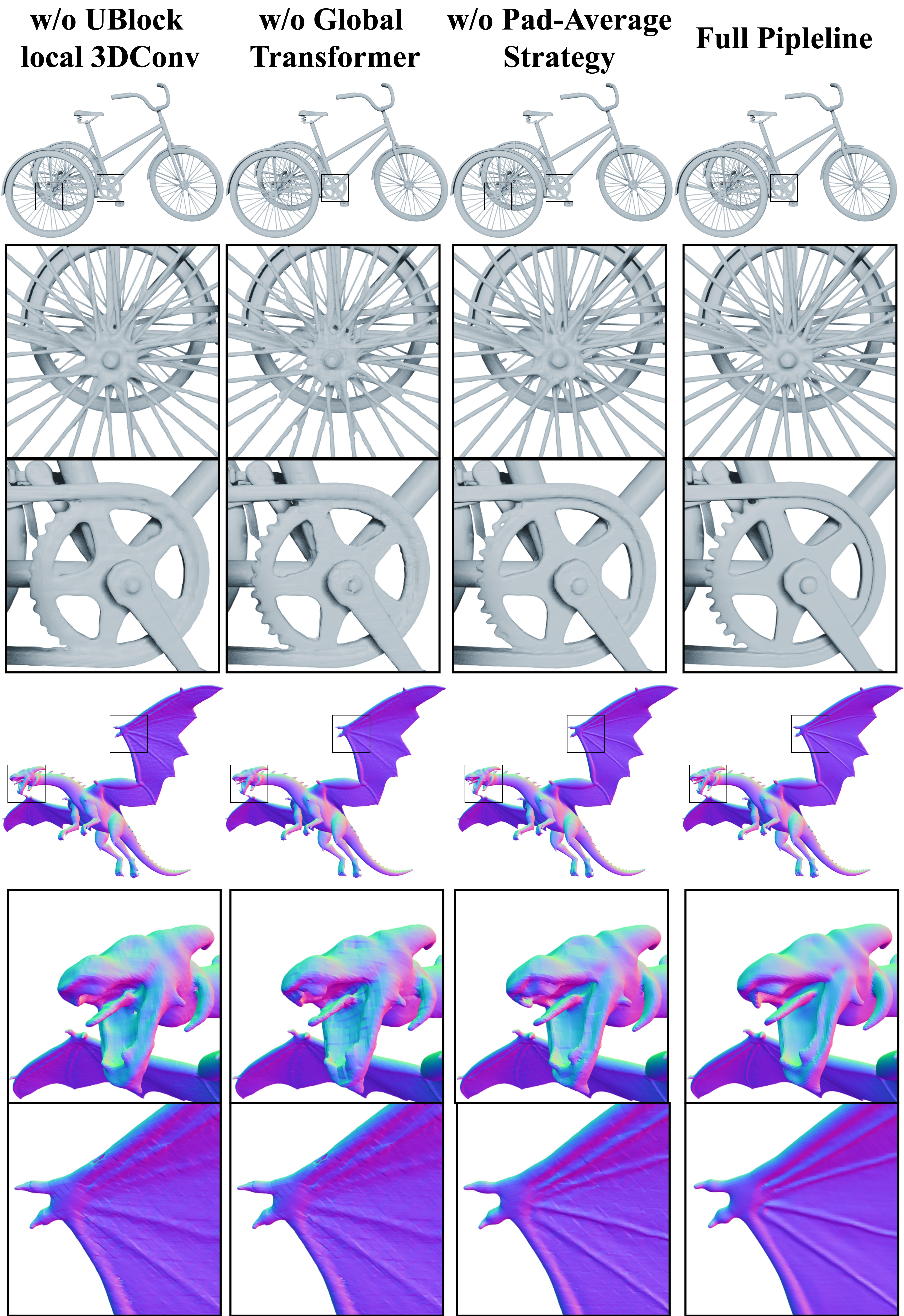}
  \caption{\textbf{Qualitative comparison of ablated models.}}
  \label{fig:ablations}
  \vspace{-4mm}
\end{figure}

We conduct ablation studies for our LoG-VAE model at $1024^3$ resolution on Toys4k dataset to analyze the impact of its key components. The results of these studies, where major design modules of our pipeline were removed, are presented qualitatively in \cref{fig:ablations} and quantitatively in \cref{tab:ablation}. 

\begin{table}
  \centering
  \caption{\textbf{Quatitative ablation study for our pipeline.} Chamfer Distance (CD, $\times 10^5$) and F1 score (F1, $\times 10^2$) are reported.}
  \resizebox{\linewidth}{!}{
  \begin{tabular}{l|cccc}
    \toprule
    Pipeline & NMSE $\downarrow$ & SNE $\downarrow$ & CD $\downarrow$ & F1(0.001) $\uparrow$ \\
    \midrule
    w/o UBlocks (local convolution) & 2.13 & 3.74 & 0.09 & 41.77  \\
    w/o Global sparse transformers  & 0.96 & 2.92 & 0.07 & 42.48  \\
    w/o Pad-Average strategy  & 0.58 & 1.90 & 0.07 & 42.55 \\
    \midrule
    Full Pipeline & \textbf{0.34} & \textbf{1.13} & \textbf{0.06} & \textbf{42.85}  \\
    \bottomrule
  \end{tabular}
  }
  \label{tab:ablation}
\end{table}

\noindent \textbf{UBlocks with Local 3D Convolutions} 
A core contribution of our work is partitioning the shape into UBlocks, which are processed by local 3D convolutions. As shown in \cref{fig:ablations}, replacing this module with a standard voxel down/up-sampling architecture (as in Direct3D-S2~\cite{wu2025direct3d} or Sparc3D~\cite{li2025sparc3d}) leads to a significant degradation in reconstruction quality. This is because our approach operates directly on full-resolution local blocks, allowing it to preserve high-frequency geometric details that are inevitably lost during the downsampling process in traditional methods.


\noindent \textbf{Global Sparse Transformers.} 
While local convolutions excel at capturing fine details, they have a limited receptive field and cannot model long-range dependencies. We introduce a global sparse Transformer to model the spatial relationships between UBlocks and ensure global structural integrity. Removing this module results in a sharp quantitative decline, as shown in \cref{tab:ablation}. Qualitatively, this manifests as visible seams and a discontinuous surface (\cref{fig:ablations}, Column 2), as the model loses its ability to enforce consistency across UBlock boundaries.


\noindent \textbf{Pad-Average Strategy.}
This strategy plays a dual role in ensuring reconstruction quality. The ``padding" step creates spatial overlap between UBlocks, providing essential context for the Transformer to learn global correlations. The subsequent ``averaging" step then smooths the transitions in these overlapping regions during reassembly, eliminating boundary artifacts. We ablate the padding value $\alpha$ in \cref{tab:Pad_value}. While performance gains begin to saturate beyond $\alpha=2$, larger values substantially increase memory consumption. Therefore, we select $\alpha=2$ as the optimal trade-off between reconstruction fidelity and computational cost.

Owing to the combined effects of these three factors, our method achieves outstanding performance in ultra-high-resolution 3D mesh reconstruction. The figures and statistics clearly demonstrate that removing any component from our pipeline results in performance degradation.


\begin{table}
  \centering
  \caption{\textbf{Different padding values in Pad-Average strategy.}}
  \begin{tabular}{c|cccc}
    \toprule
    Padding Value & $\alpha=0$ & $\alpha=1$ & $\alpha=2$ & $\alpha=3$ \\
    \midrule
    NMSE $\downarrow$ & 0.58 & 0.40 & 0.34 & \textbf{0.33} \\
    \midrule
    SNE $\downarrow$ & 1.90 & 1.24 & \textbf{1.13} & \textbf{1.13} \\
    \bottomrule
  \end{tabular}
  \vspace{-2mm}
  \label{tab:Pad_value}
\end{table}

\section{Limitations and Future Work}
\label{Limitations}
While LoG-VAE demonstrates superior geometric fidelity in detail preservation, we identify several limitations. First, our current framework focuses exclusively on geometry and lacks an explicit mechanism for texture generation. 
Second, volumetric reconstruction at high resolution of  $2048^3$ incurs heavy computational overhead for MarchingCubes~\cite{lorensen1998marching}. These limitations could be addressed by integrating spectral texture fields, employing hybrid training pipelines, and adopting GPU-accelerated isosurface extraction.
\section{Conclusion}
\label{Conclusions}
In this work, we present a novel framework for 3D Variational Auto-Encoders based on unsigned distance fields (UDFs), addressing fundamental limitations of conventional representations such as signed distance functions and point clouds. By introducing UBlocks partitioning, pad-average strategy, and a local-to-global (LoG) VAE architecture, our method enables high-fidelity reconstruction and generation of complex 3D shapes. Our method successfully scales to ultra-high resolutions of up to $2048^3$, achieving state-of-the-art geometric accuracy and surface smoothness. We believe LoG3D not only sets a new standard for high-resolution 3D reconstruction but also opens new avenues for creating detailed 3D contents for real-world applications.

{
    \small
    \bibliographystyle{ieeenat_fullname}
    \bibliography{main}
}


\end{document}